\documentclass[conference]{IEEEtran}
\usepackage{cite}
\usepackage{subcaption}
\usepackage{amsmath,amssymb,amsfonts}
\usepackage{algorithmic}
\usepackage{graphicx}
\usepackage{textcomp}
\usepackage{xcolor}

\begin{document}

\title{Automated Weld Seam Recognition and 3D Mapping for Robotic Post Processing Using Photogrammetry and Semantic Segmentation}

\author{ \IEEEauthorblockN{ Augustin Raju\IEEEauthorrefmark{1}, Abilash Madavath\IEEEauthorrefmark{1}, Chandra Yuvesh Aubeeluck\IEEEauthorrefmark{1} } \IEEEauthorblockN{ Nicolas Pyschny\IEEEauthorrefmark{1}, Felix Hackelöer\IEEEauthorrefmark{1}, Florian Zwanzig\IEEEauthorrefmark{1} } \IEEEauthorblockA{ \IEEEauthorrefmark{1}TH Köln - University of Applied Sciences, Germany } }
\maketitle

\begin{abstract}
Accurate identification of weld seam geometries is essential for automated robotic post processing operations such as grinding, finishing, and inspection. For large workpieces, complete surface scanning using high precision laser scanners or structured light sensors can be time consuming and often generates substantial amount of data that are not relevant. This paper presents an experimental vision based pipeline for the approximate localization of weld seams. This serves as a preliminary stage before high precision measurement. The proposed approach aims to reduce the overall scanning effort and data acquisition efficiency. The proposed method includes capturing images of the workpiece from multiple viewpoints, identifying weld seams from the images using semantic segmentation, reconstructing the workpiece using photogrammetry, and projection of identified weld seams into the reconstructed model. 
\end{abstract} 
\section{Introduction}

Welded components often require post processing operations such as grinding, smoothing, finishing, or inspection after the welding process. In many industrial environments, these tasks are still performed manually or with limited automation. For robotic systems to perform these post processing operations, accurate weld seam geometry in 3D space is required.

Many studies have investigated identification of weld seams from 3D point clouds and meshes captured using laser or structured light cameras \cite{chen2025recent, zhong2026intelligent, liu2024novel}. While these approaches can provide accurate geometric information, they are often suited for small components mainly due to scanning speed and measurement volume. An alternative under explored approach for gathering 3D data is by using photogrammetry. Very few works have explored the use of photogrammetry for weld seam localization and 3D reconstruction \cite{ona2022weldmap, viteri2025industrial}. They were limited to small measurement volumes.

For large workpieces, complete high resolution scanning can become inefficient as some data may not be relevant for further post processing. It is beneficial to have a preliminary stage to guide subsequent high precision measurements as it reduce scanning effort by limiting detailed measurements to regions that contains weld seams. Previous research has explored the use of low resolution depth cameras for this purpose \cite{li2023guidance}. However, their limited spatial resolution often prevents reliable detection of narrow or complex weld geometries.

Motivated by these challenges, this work investigates an alternative weld seam localization approach based on images captured from multiple viewpoints using a standard smartphone camera. By combining image based weld detection with photogrammetric 3D reconstruction, the proposed method aims to provide an efficient first stage localization process that supports targeted high precision scanning and subsequent robotic post processing operations.

\section{Methodology}

\begin{figure}[t] 
\centering 
\includegraphics[width=0.75\columnwidth]{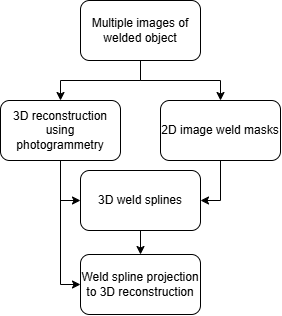} 
\caption{Overview of the proposed workflow. Multiple images of a welded object are used for photogrammetric 3D reconstruction and weld seam segmentation. The resulting information is combined to generate and project 3D weld splines onto the reconstructed object geometry.} 
\label{fig:flowchart} 
\end{figure}

Figure~\ref{fig:flowchart} illustrates the overall methodology proposed in this work. The workflow consists of two parallel processing branches. The first branch performs photogrammetric reconstruction of the welded component using multiple overlapping images, producing a 3D representation of the object. The second branch uses the same image set for weld seam detection, resulting in 2D weld masks. Information from both branches is combined to reconstruct 3D weld splines, which are subsequently projected onto the reconstructed object geometry to obtain a spatial representation of the weld seam.

\subsection{Data Acquisition and 3D Reconstruction}
The welded workpiece is surrounded with AprilTag markers. The marker dimensions are already known and they are used for scaling and frame matching. Images of the welded workpieces are captured from multiple viewpoints with markers visible. Each of the images have some overlapping features with each other to support photogrammetric reconstruction. 

The acquired image set is then processed using RealityScan, a photogrammetry based tool to generate a 3D reconstruction of the workpiece. Additionally, it generated the camera poses, intrinsic parameters and depth map of the images taken.

\subsection{Weld Seam Segmentation \& Centerline Extraction}

Weld seam recognition is performed using a SegFormer based semantic segmentation network. The problem is formulated as a binary classification task consisting of weld and background classes. The training data consisted of 388 images collected from online public datasets and the surrounding environment. After training, the network generates binary masks that identify weld seam regions within input images.

The predicted weld masks are processed to obtain a thinner path representation of the seam. First, small and noisy mask regions are filtered. Connected weld components are identified, and principal component analysis is applied to estimate the dominant orientation of each weld region. The mask is then divided into local slices along the estimated direction, and center points are computed for each slice. The resulting point sequence forms a spline like representation of the weld seam in image space.

\subsection{3D Projection and Multiview Fusion}

The extracted spline points are projected into 3D space using the depth maps and camera calibration data obtained during reconstruction. Each image space point is first converted into camera coordinates and subsequently transformed into the global reconstruction coordinate system using the corresponding camera pose.

Since the same weld seam may appear in multiple images, all reconstructed weld points are transformed into a common coordinate frame and merged. Duplicate observations are removed and outlier filtering is applied to obtain a consolidated 3D weld seam representation.

\begin{figure}[t] 
\centering 
\includegraphics[width=\columnwidth]{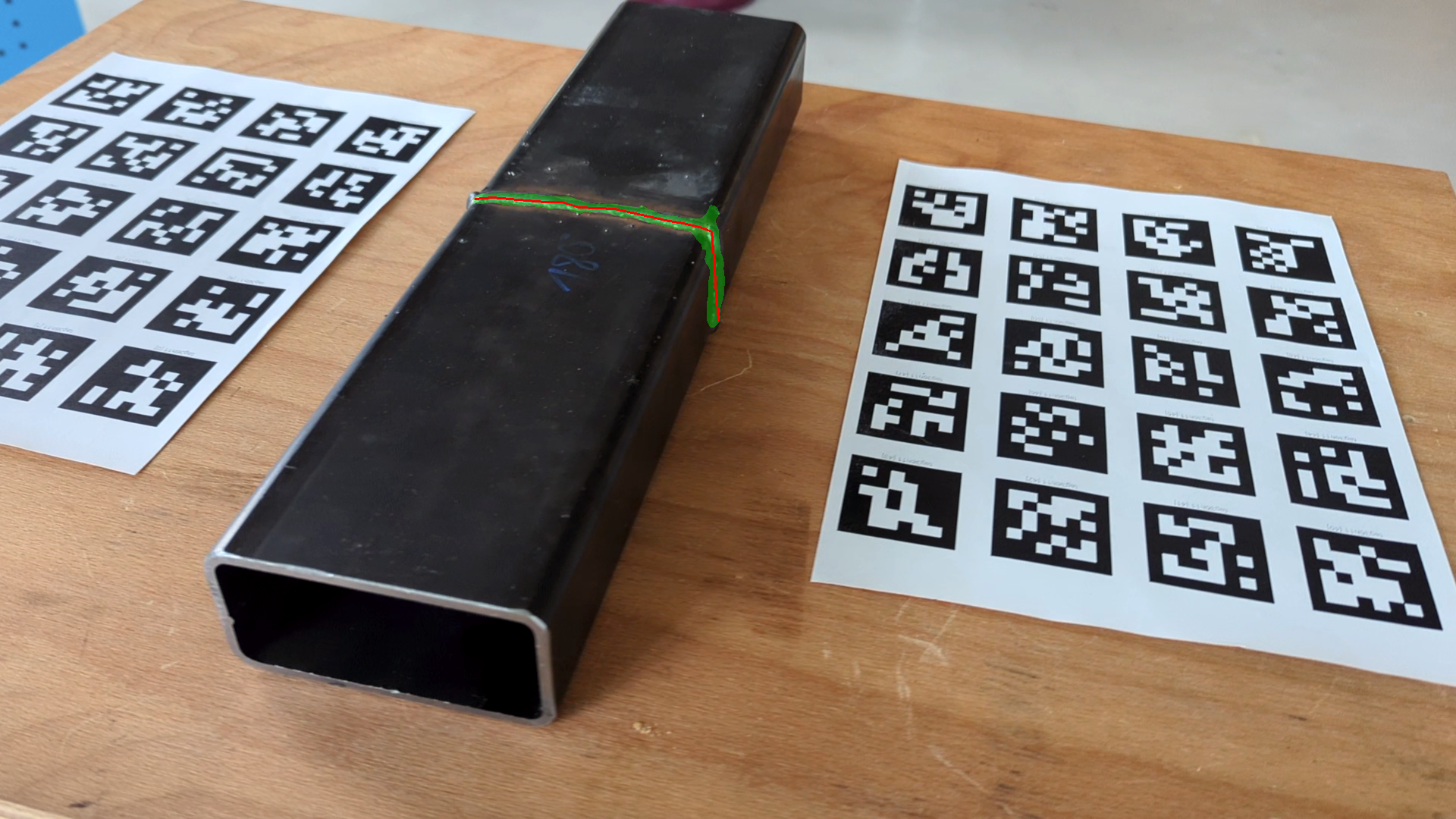} 
\caption{An image captured that shows the experimental setup. Weld mask is shown as green overlay. Centerline marking is shown as red line overlay.} 
\label{fig:spline} 
\end{figure}

\section{Experimental Setup}

To evaluate the feasibility of the proposed approach, a simple welded test workpiece was selected. The workpiece consists of two identical rectangular hollow steel tube sections joined by a continuous circumferential butt weld, with overall outer dimensions of 40 cm × 10 cm × 5 cm (L × W × H). To provide geometric reference features for measurement and localization, two AprilTag marker sheets were placed adjacent to the workpiece. The center to center distance between adjacent markers on each sheet was 5.2 cm.

A total of 68 images each with a resolution of 1920 * 1080 pixels were acquired using a smartphone camera. Images were collected from all accessible sides of the workpiece to obtain comprehensive visual coverage of both the workpiece geometry and the weld seam.

The acquired image set was subsequently used for both photogrammetric reconstruction and weld seam detection. Figure \ref{fig:spline} illustrates the experimental setup and shows an example of a captured image.

\section{Results and Discussion}

As this study represents an initial proof of concept investigation, the primary objective was to evaluate the overall feasibility of the proposed workflow rather than to rigourously quantify the accuracy of each individual processing stage. 

\begin{figure}[t] 
\centering 
\includegraphics[width=\columnwidth]{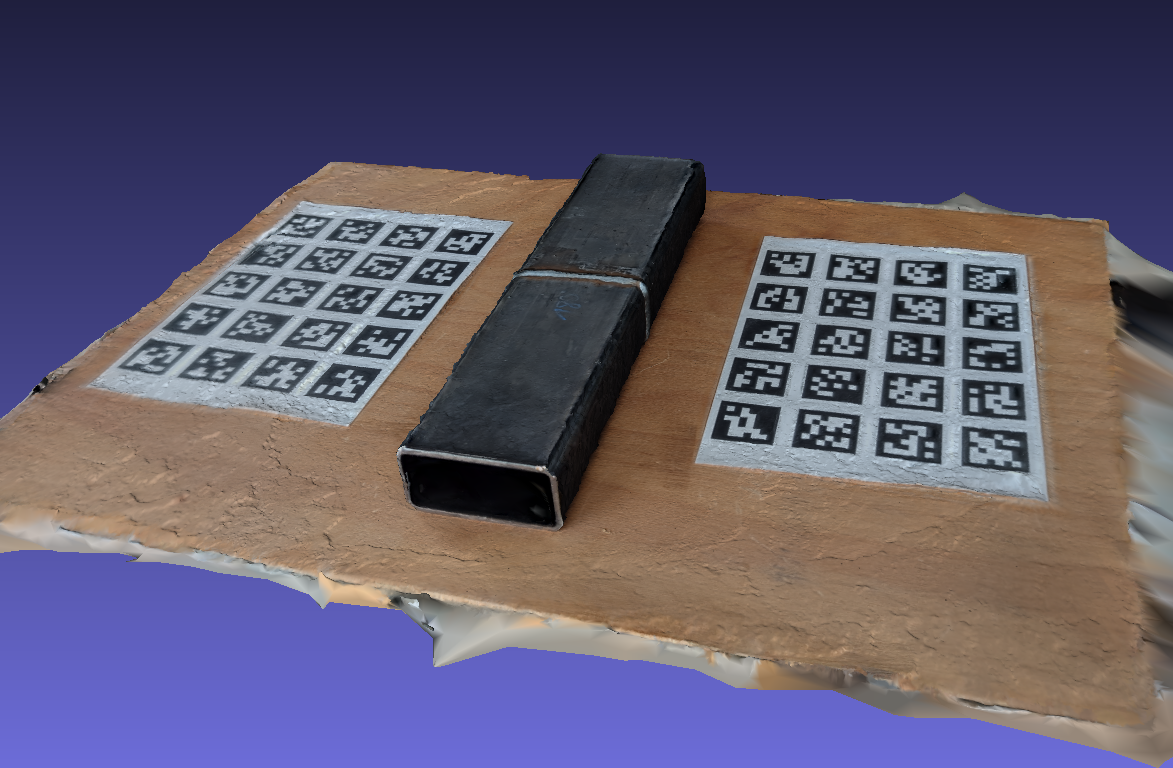} 
\caption{3D reconstruction of the workpiece using photogrammetry} 
\label{fig:resultA} 
\end{figure}

The photogrammetry based reconstruction successfully generated a 3D point cloud with a scale consistent with the physical dimensions of the workpiece. It is shown in Figure \ref{fig:resultA}. The reconstructed model had a dimensional error of approximately 3 mm when compared to the actual outer dimensions of the workpiece. The overall geometry of the workpiece was reconstructed, although minor local deviations and surface mismatches were observed. Since the proposed method is intended to provide only an approximate localization of weld seams for guiding subsequent high precision measurements, these reconstruction inaccuracies are considered acceptable. 

The custom trained SegFormer model was able to identify and segment the weld seam in the majority of the captured images. Missed detections occurred in a small number of cases, primarily due to variations in lighting conditions and surface reflections. Additionally, some no weld regions were occasionally misclassified as welds, resulting in the generation of weld spline points in areas where no actual weld was present.

\begin{figure}[t] 
\centering 
\includegraphics[width=\columnwidth]{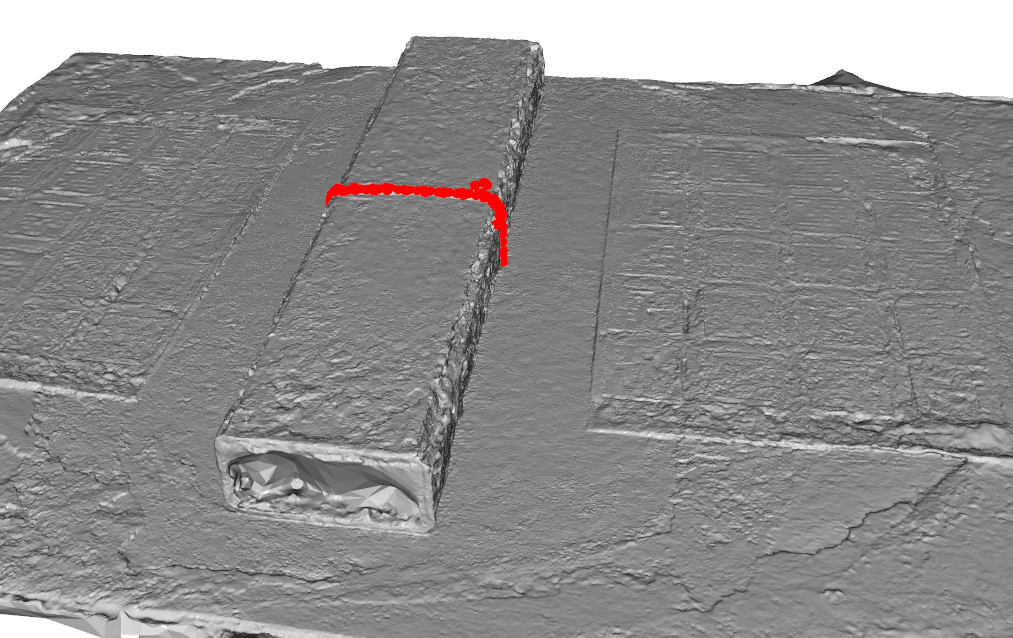} 
\caption{Projection of weld spline points into the 3D reconstruction. Weld spline points are shown as red dots.} 
\label{fig:resultB} 
\end{figure}

The extracted weld spline points were subsequently projected onto the reconstructed 3D geometry using the corresponding depth and camera pose information. It is shown in Figure \ref{fig:resultB}. The resulting spatial representation accurately indicated the approximate location and extent of the weld seam on the workpiece. To evaluate the localization performance, the projected weld spline points were compared with a manually annotated reference weld line, yielding a root mean square error (RMSE) of 4.2 mm.

The experimental results demonstrate the feasibility of combining image based weld seam detection with photogrammetric reconstruction for coarse weld localization. Despite minor reconstruction inaccuracies and occasional segmentation errors, the proposed workflow successfully generated a spatial weld seam map with millimeter level localization accuracy. Such a map can serve as a guide for subsequent high precision inspection or measurement procedures, reducing overall scanning effort and restricting detailed data acquisition to regions of interest.

\section{Limitations \& Future Work}

There exists many limitations that needs to be addressed. The segmentation model is currently trained on a limited dataset, making performance sensitive to lighting conditions, reflections, image quality, and variations in weld appearance. Similar challenges affect photogrammetric reconstruction because it relies on image quality and surface texture.

Future work will focus on expanding the training dataset, applying stronger augmentation strategies, improving 3D weld path refinement, and estimating local surface normals and tool orientations. The long term objective is to generate executable robot trajectories for automated weld grinding, finishing, and inspection.

\section{Conclusion}

This paper presented an experimental pipeline for weld seam localization by combining semantic segmentation, photogrammetric reconstruction, and 3D weld mapping. Images acquired from multiple viewpoints were used to detect weld seams and reconstruct the workpiece geometry, enabling the identified weld regions to be projected into a common 3D coordinate system.

Initial results show that a custom trained SegFormer model can identify weld seam regions in captured images and that these detections can be converted into approximate 3D weld paths using depth maps and camera poses.

The main contribution of this work is an integrated experimental pipeline that connects weld seam segmentation, mask based centerline extraction, depth assisted 3D projection, and multi view weld point fusion. Unlike approaches that focus only on 2D weld detection, the proposed workflow aims to localize the weld seam directly on the reconstructed object geometry. This makes it suitable as a foundation for future robotic grinding, finishing, or inspection applications.

\section*{Acknowledgment}

This work is supported through the InnoFaktur project by the European Regional
Development Fund under grant EFRE-20500002.\\
The authors thank August Rüggeberg GmbH \& Co.~KG (PFERD) for domain access and support.

\bibliographystyle{plain} 
\bibliography{references}

\end{document}